\documentclass[10pt,twocolumn,letterpaper]{article}

\usepackage{cvpr}
\usepackage{times}
\usepackage{epsfig}
\usepackage{graphicx}
\usepackage{amsmath}
\usepackage{amssymb}
\usepackage{adjustbox}
\usepackage{boldline}
\usepackage[numbers, sort]{natbib}
\usepackage{multirow}
\usepackage{afterpage}
\usepackage{color}

\usepackage[pagebackref=true,breaklinks=true,letterpaper=true,colorlinks,bookmarks=false]{hyperref}

\cvprfinalcopy 

\def\cvprPaperID{2660} 

\ifcvprfinal\fi
\begin{document}

\title{Deep Face Super-Resolution with Iterative Collaboration between Attentive Recovery and Landmark Estimation}

\author{
Cheng Ma$^{1,2,3}$, Zhenyu Jiang$^{1}$, Yongming Rao$^{1,2,3}$, Jiwen Lu$^{1,2,3}\thanks{Corresponding author}$\ , Jie Zhou$^{1,2,3,4}$\\
{$^1$Department of Automation, Tsinghua University, China}\\
{$^2$State Key Lab of Intelligent Technologies and Systems, China}\\
{$^3$Beijing National Research Center for Information Science and Technology, China}\\
{$^4$Tsinghua Shenzhen International Graduate School, Tsinghua University, China} \\
{\tt\small macheng17@mails.tsinghua.edu.cn; jiangzhe16@mails.tsinghua.edu.cn} \\ {\tt\small raoyongming95@gmail.com; lujiwen@tsinghua.edu.cn; jzhou@tsinghua.edu.cn}\\
}

\maketitle
\thispagestyle{empty}

\begin{abstract}
   Recent works based on deep learning and facial priors have succeeded in super-resolving severely degraded facial images. However, the prior knowledge is not fully exploited in existing methods, since facial priors such as landmark and component maps are always estimated by low-resolution or coarsely super-resolved images, which may be inaccurate and thus affect the recovery performance. 
   In this paper, we propose a deep face super-resolution (FSR) method with iterative collaboration between two recurrent networks which focus on facial image recovery and landmark estimation respectively. In each recurrent step, the recovery branch utilizes the prior knowledge of landmarks to yield higher-quality images which facilitate more accurate landmark estimation in turn. 
   Therefore, the iterative information interaction between two processes boosts the performance of each other progressively. 
   Moreover, a new attentive fusion module is designed to strengthen the guidance of landmark maps, where facial components are generated individually and aggregated  attentively for better restoration. 
   Quantitative and qualitative experimental results show the proposed method significantly outperforms state-of-the-art FSR methods in recovering high-quality face images. \footnote{Code: \href{https://github.com/Maclory/Deep-Iterative-Collaboration}{https://github.com/Maclory/Deep-Iterative-Collaboration}}
\end{abstract}

\section{Introduction}

\begin{figure}
\centering
\includegraphics[width=\linewidth]{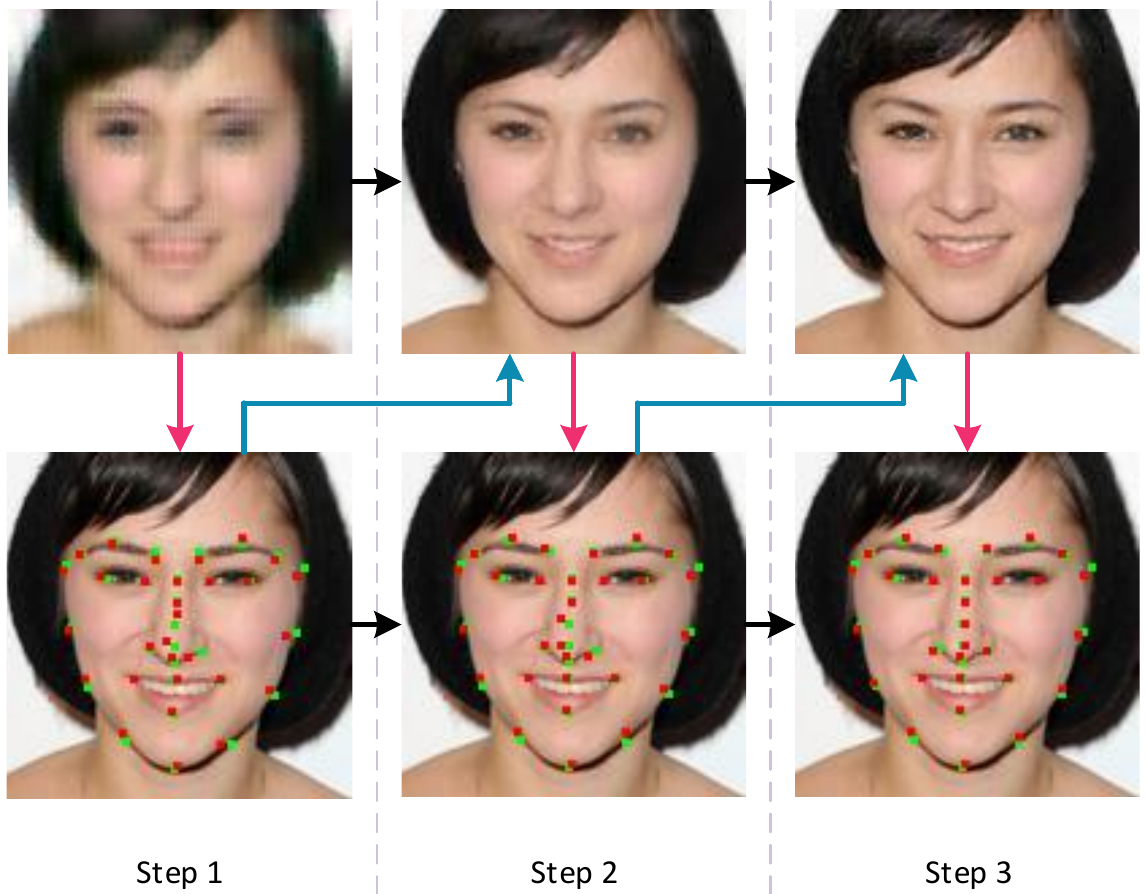}
\vspace{-3mm}
\caption{Data flow of the proposed method. FSR outputs in different steps are shown in the top row while the detected facial landmarks are displayed on HR images accordingly in the bottom row. The pink arrows denote the face alignment process while the blue ones denote the face recovery process with the attentive fusion of landmarks. 
The black arrows represent the recurrent schemes in two branches. Through our framework, the quality of SR images becomes better progressively and the estimated landmarks (red) also get closer to the ground-truth (green).}
\label{fig:head}
\vspace{-5mm}
\end{figure}

In recent years, face super-resolution (FSR), also known as face hallucination, has attracted much attention of the computer vision community. FSR aims to restore high-resolution (HR) face images from the low-resolution (LR) counterparts, which plays an important role in many applications such as video surveillance and face enhancement. Moreover, facial analysis techniques including face recognition and face alignment can also benefit a lot from the quality improvement brought by FSR. 

FSR is a special case of the task of single image super-resolution (SISR)~\cite{RCAN,shi2016real,SFTGAN,EnhanceNet,wang2018esrgan}, which is  a challenging problem since it is highly ill-posed due to the ambiguity of the super-resolved pixels. Compared to SISR, FSR only considers facial images instead of arbitrary scenes. Therefore, the specific facial configuration can be strong prior knowledge for the generation, so that global structures and local details can be recovered accordingly. Hence FSR methods perform better than SISR on higher upscaling factors (e.g., $8\times$). A number of methods for face super-resolution~\cite{ma2010hallucinating,yang2013structured,jia2008generalized,jiang2014face,huang2010super,chakrabarti2007super,liu2007face,wang2014comprehensive,jung2011position} have been proposed recently. Furthermore, the advent of deep learning techniques has greatly boosted the performance of face hallucination because of the powerful generative ability of deep convolutional neural networks (DCNNs). 

Facial priors have been utilized in existing FSR methods. Dense correspondence field is used in~\cite{zhu2016deep} to capture the information of face spatial configuration. 
Facial component heatmaps are predicted in~\cite{yu2018face} to provide localizations of facial components for improving the SR quality. An end-to-end trained network~\cite{chen2018fsrnet} introduces facial landmark heatmaps and parsing maps simultaneously to boost the recovery performance. However, there are some limitations with such methods. On the one hand, they have difficulty in estimating accurate prior information for the reason that the localization and alignment processes are applied on LR input images or coarse SR images which are of low quality and far from final results. Hence given inexact priors, the guidance for SR may be erroneous. On the other hand, most methods just optimize the recovery and prior prediction as a problem of multi-task learning and incorporate the prior information by a simple concatenation operation. However, such guidance is not direct and clear enough since the structural variations of different components may not be fully captured and exploited. Therefore, more powerful schemes to utilize facial priors should be explored. 

In this paper, we propose a deep iterative collaboration method for face super-resolution to mitigate the above issues. Firstly, we design a new framework including two branches, one for face recovery and the other for landmark estimation. Different from previous methods,
we let the face SR and alignment processes facilitate each other progressively. The idea is inspired by the fact that the SR branch can generate high-fidelity face images with the guidance of accurate landmark maps and the alignment branch also benefits a lot from high-quality input images. 
To achieve this goal, we build a recurrent architecture instead of very deep generative models for SR while designing a recurrent hourglass network for face alignment, rather than conventional stacked hourglass networks~\cite{newell2016stacked}. In each recurrent step, previous outputs of each branch are fed into the other branch in the following step, so that both branches collaborate with each other for better performance. 
Moreover, the feedback schemes implemented in two branches both increase the efficiency of the whole framework. Secondly, we propose a new attentive fusion module to integrate the landmark information instead of the concatenation operation. Specifically, we utilize the estimated landmark maps to generate multiple attention maps, each of which reveals the geometric configuration of one facial key component. Benefiting from the component-specific attention mechanism, features for each component can be extracted individually, which can be easily accomplished by group convolutions. 
Experimental results on two popular benchmark datasets, CelebA~\cite{liu2015deep} and Helen~\cite{le2012interactive}, demonstrate the superiority of our method in super-resolving high-quality face images over state-of-the-art FSR methods.

\section{Related Work}
\textbf{Face Super-Resolution}:
Recently, deep learning based methods have achieved remarkable progress in various computer vision tasks including face super-resolution. Yu \etal~\cite{yu2016ultra} introduce a deep discriminative generative network that can super-resolve very low face images. Huang \etal~\cite{huang2017wavelet} turn to wavelet domain and propose a network that predicts wavelet coefficients of HR images. Besides, Yu \etal\cite{yu2018super} embed attributes in the process of face super-resolution. 
Zhang \etal~\cite{zhang2018super} introduce a super-identity loss to measure the identity difference. 
Some face SR methods also divide the solution into global and local parts. Tuzel \etal~\cite{tuzel2016global} design a network that contains two sub-networks: the first one reconstructs face images based on global constraints while the second one enhances local details. Cao \etal~\cite{cao2017attention} propose to use reinforcement learning to specify attended regions and use a local enhancement network for recovery sequentially. 

Since face hallucination is a domain-specific task, facial priors are utilized in some FSR methods. Yu \etal~\cite{yu2018face} concatenate facial component heatmaps with features in the middle of the network. Chen \etal~\cite{chen2018fsrnet} concatenate facial landmark heatmaps and parsing maps with features. Kim \etal~\cite{kim2019progressive} design a facial attention loss based on facial landmark heatmaps and use it to train a progressive generator.  
Zhu \etal~\cite{zhu2016deep} propose a deep bi-network which conducts face hallucination and face correspondence alternatively to refine both processes progressively. However, the architecture of the cascaded framework is redundant and inflexible, restricting the efficiency of the model. Moreover, the lack of ability to estimate accurate dense corresponding fields may also lead to severe distortions.

\begin{figure*}
\centering
\includegraphics[width=0.95\linewidth]{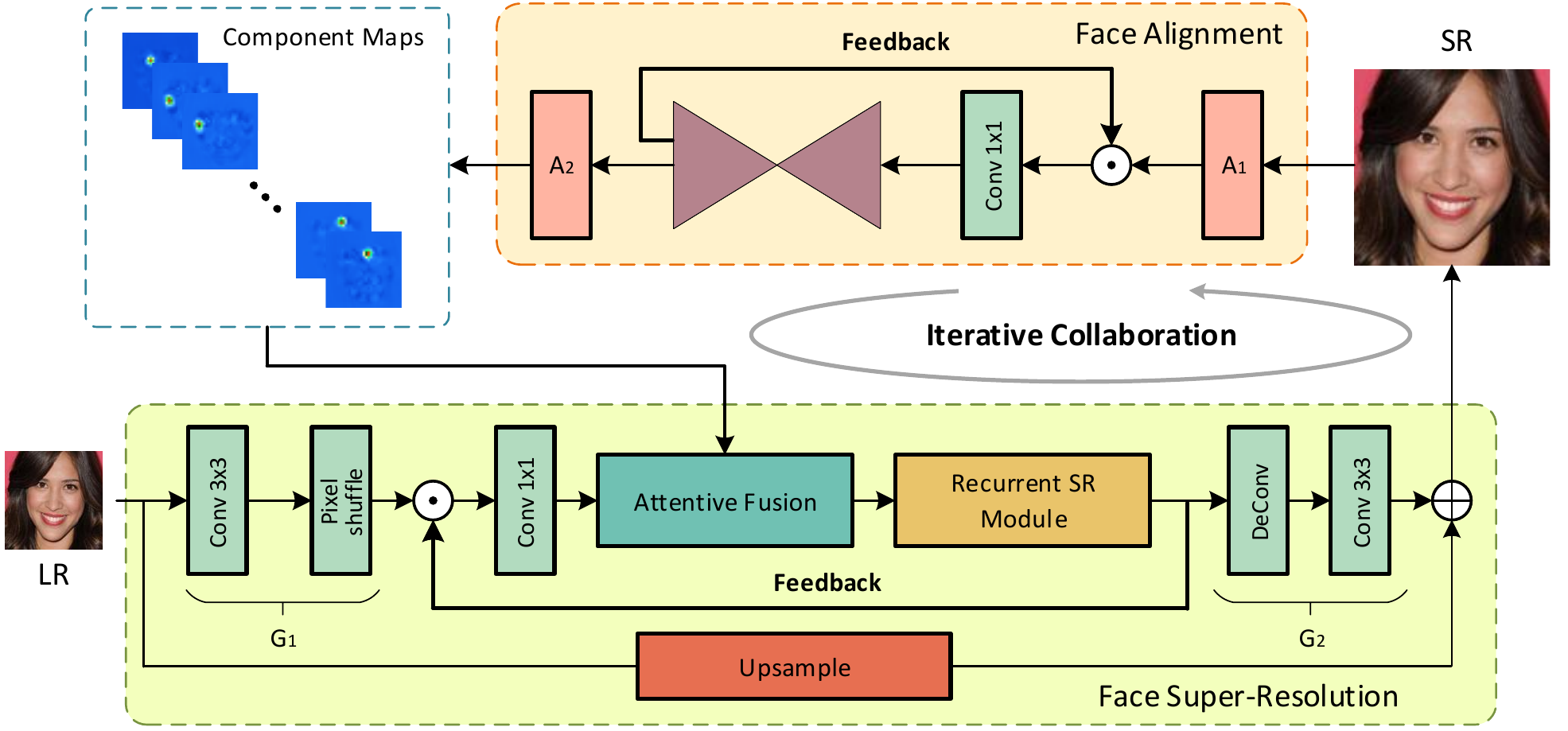}
\caption{Overall framework of the proposed deep iterative collaboration method. The architecture is comprised of two branches, a recurrent SR branch and a recurrent alignment branch. Two branches collaborate with each other and obtain better SR images and more accurate landmarks step by step. ``$\odot$'' and ``$\oplus$'' denote concatenation and addition respectively. }
\vspace{-3mm}
\label{fig:framework}
\end{figure*}

\textbf{Single Image Super-Resolution}
As a pioneer of using deep networks in single image super-resolution (SISR), Dong \etal~\cite{dong2014learning} propose SRCNN to learn a mapping from bicubic-interpolated images to HR images. Kim \etal~\cite{kim2016accurate} propose VDSR by using a 20-layer VGG-net~\cite{simonyan2014very} to learn the residual of LR and HR images.  
Methods mentioned above mainly focus on PSNR and SSIM. Their results are mostly blurry. 
Recently, perceptual quality of SR images is drawing more and more attention. 
SRGAN \etal~\cite{ledig2017photo} is the first to generate photo-realistic images with the adversarial loss and the perceptual loss~\cite{johnson2016perceptual}.
Rad \etal~\cite{rad2019srobb} extend the perceptual loss with a targeted perceptual loss. 

Recently, recurrent networks have also been utilized for SISR.
Kim \etal~\cite{kim2016deeply} propose DRCN, a deep recursive CNN, and obtain outstanding performance compared to previous work. Tai \etal~\cite{tai2017image} use residual units to build deep and concise networks with recursive blocks.  Zhang \etal~\cite{zhang2018residual} follow the idea of DenseNet~\cite{huang2017densely} and design a residual dense block to fuse hierarchical features.  Han \etal~\cite{han2018image} design a dual-state recurrent network that exploits LR and HR signals jointly. Li \etal~\cite{li2019feedback} introduce a new feedback block where features are iteratively upsampled and downsampled. 
While recursive networks promote the development of SISR, few methods have employed their generative power in face super-resolution. Hence it remains an attractive direction to exploit the potential ability of recurrent mechanisms for FSR.

\section{Approach}

In face super-resolution, we aim to recover the facial details of input LR face images $I^{LR}$ and get the SR results $I^{SR}$. We design a deep iterative collaboration network which estimates high-quality SR images and landmark maps iteratively and progressively with the input LR images. In order to enhance the collaboration between the SR and alignment processes, we design a novel attentive fusion module that integrates two sources of information effectively. 
Finally, we apply an adversarial loss to supervise the training of the framework and produce enhanced SR faces with high-fidelity details. 

\subsection{Deep Iterative Collaboration}

\begin{figure*}
\centering
\includegraphics[width=\linewidth]{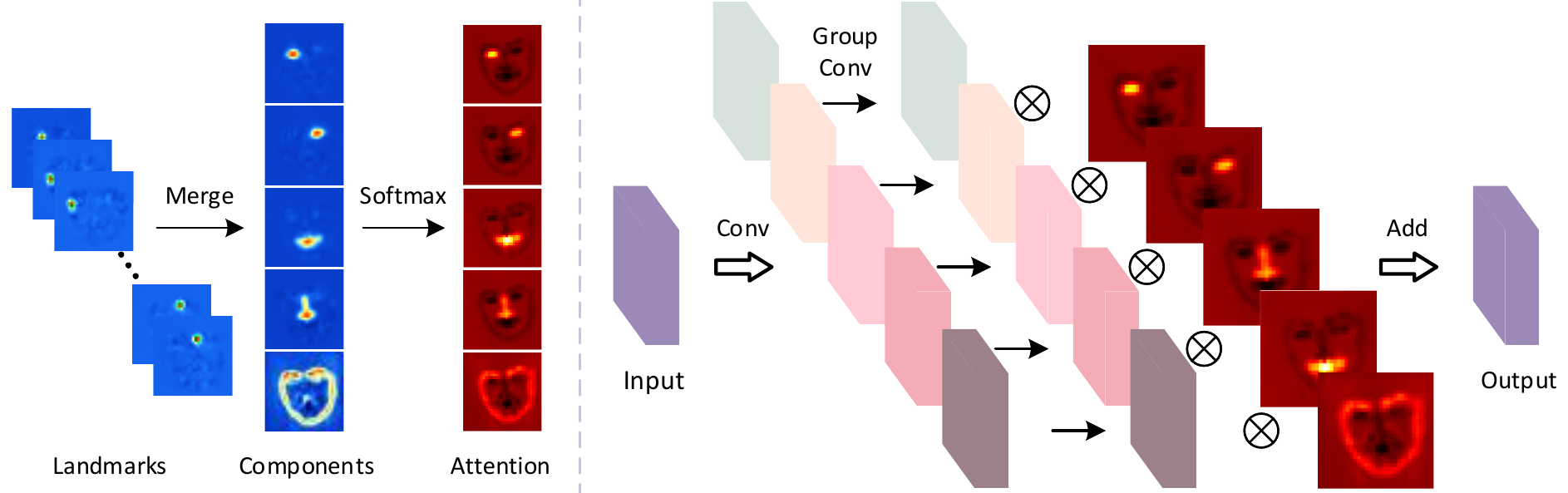}
\vspace{-2mm}
\caption{The left part illustrates the method to extract attention maps from landmark maps. The right part shows the flowchart of the attentive fusion module. 
The input feature is expanded by a convolutional layer. Then component-specific features are extracted by a series of group convolutional layers under the guidance of attention maps.  We multiply (``$\otimes$'') the features with the attention maps which are broadcast through the channel dimension. Finally, weighted features are added together to form the output.}
\label{fig:fusion}
\vspace{-2mm}
\end{figure*}

Given an LR face image $I^{LR}$, facial landmarks are important for the recovery procedure. However, prior estimation via LR faces is unreliable since a lot of details are missing. Such information may provide inaccurate guidance to SR effects. Therefore, our method alleviates this issue by an iterative collaborative scheme as shown in Figure~\ref{fig:framework}. In this framework, face recovery and landmark localization are performed simultaneously and recursively. We can get better SR images by accurate landmark maps as landmarks are estimated more correctly if the input faces have higher quality. Both processes can enhance each other and achieve better performance progressively. Finally, we can get accurate SR results and landmark heatmaps with enough steps. 

The recurrent SR branch $G$ consists of a low-resolution feature extractor $G_1$, a recursive block $G_R$ and high-resolution generation layers $G_2$. $G_R$ includes an attentive fusion module and a recurrent SR module. 
Similar to the SR branch, the recurrent alignment branch includes a pre-processing block $A_1$, a recursive hourglass block $A_R$ and a post-processing block $A_2$. 
For the $n$th step where $n=1, \cdots, N$, the SR branch recovers SR images $I_n^{SR}$ by using the alignment results and the feedback information from the previous step $n-1$, denoted as $L_{n-1}$ and $f_{n-1}^{G_R}$, respectively. Besides, LR inputs are also important in each step. Hence LR features extracted by $G_1$ are also fed into the recursive block. 
Therefore, the face SR process can be formulated by:
\begin{eqnarray}
f_n^{G_R} &=& G_R(G_1(I^{LR}), f_{n-1}^{G_R}, L_{n-1}), \\
I_n^{SR} &=& G_2(f_n^{G_R})+U(I^{LR}),
\end{eqnarray}
where $U$ denotes an upsampling operation. 
Similarly, the face alignment branch utilizes the recurrent features from the previous step $f_{n-1}^{A_R}$ and the SR features extracted by $A_1$ from the SR images $I_{n}^{SR}$ as the guidance for estimating landmarks more accurately, as follows: 
\begin{eqnarray}
f_n^{A_R} &=& A_R(A_1(I_{n}^{SR}), f_{n-1}^{A_R}),\\
L_n &=& A_2(f_n^{A_R}).
\end{eqnarray}
After $N$ steps, we get $\{I^{SR}_n\}^N_{n=1}$ and $\{L_{n}\}_{n=1}^N$ where outputs become more satisfactory as $n$ increases. 
In the beginning, there is no recurrent feature and landmark map from the previous step. 
Therefore, we use an extra similar SR module which takes only the LR features as input before the first step to get $f_0^{G_R}$ as an initialization for the following steps. Meanwhile, we make $f_0^{A_R}=A_1(I_1^{SR})$ to initialize the face alignment branch.  

For the purpose of achieving more powerful optimization, we impose loss functions to each output of $N$ steps. By this means, the SR and alignment are strengthened in every step and the inaccurate factors are corrected gradually by mutual supervision. Here, the pixel-wise loss functions are defined as follows:
\begin{eqnarray}
\mathcal{L}^{Pixel} &=& \mathbb{E}\left[\frac{1}{N}\sum^{N}_{n=1}\|I^{HR}-I^{SR}_n\|_2^2\right], \\
\mathcal{L}^{Align} &=& \mathbb{E}\left[\frac{1}{N}\sum^{N}_{n=1}\|L^{HR}-L_n\|_2^2\right],
\end{eqnarray}
where $\mathcal{L}^{Pixel}$ and $\mathcal{L}^{Align}$ are the loss functions for the face SR and landmark estimation, respectively. $I^{HR}$ and $L^{HR}$ are the ground-truth HR images and landmark heatmaps. We use SR images in the last step as the final outputs, which can be formulated as $I^{SR}=G(I^{LR})$.

\subsection{Attentive Fusion Module}

In existing methods, straight-forward ways of utilizing facial prior knowledge are to concatenate facial priors with SR features and treat the whole optimization procedure as a problem of multi-task learning. However, facial structures may not be fully exploited since features of different facial parts are usually extracted by a shared network. 
Hence the specific structural configuration priors existing in different facial components may be neglected by the networks. Therefore, different facial parts should be recovered separately for better performance. \cite{cao2017attention} has exploited the global interdependency of facial parts by reinforcement learning. 
However, the sequential patch reconstruction cannot utilize facial priors explicitly and efficiently, which also limits the specialized generation for different facial components.

Differently, we achieve the above goals by a new structure-aware attentive fusion module so as to make full use of the guidance of landmarks $L$.
We assume each landmark heatmap has $K$ channels indicating the locations of $K$ landmarks. The landmarks can be grouped into $P$ subsets, belonging to facial components including left eye, right eye, nose, mouth and jawline. Channels in each group are added together to form the heatmap for the corresponding facial component, denoted as $\{C_p\}_{p=1}^P$ and shown in Figure~\ref{fig:fusion}. 
The reason to do so rather than directly fuse the learned landmarks is in two aspects: 
(1) We explicitly highlight the local structure of  each facial parts to perform differential recovery; 
(2) The number of channels is largely reduced by the grouping process so as to improve the efficiency of the framework. 
Then we can compute $P$ corresponding attention maps by the softmax function along the channel dimension of these heatmaps, as below:
\begin{eqnarray}
M_p(x,y) &=& \frac{e^{C_p(x,y)}}{\sum_{j=1}^P e^{C_j(x,y)}},
\end{eqnarray}
where $(x,y)$ represent the spatial coordinates of attention map $M_p$. Instead of using multiple models for different facial components, we apply group convolutions to generate individual features $f_p$. The flow chart is depicted as Figure~\ref{fig:fusion}. In order to make each group of convolutions concentrate on the corresponding parts, we define an attentive fusion as:
\begin{eqnarray}
f_{Fusion} &=& \sum_{p=1}^P M_p \cdot f_p,
\end{eqnarray}
where $f_{Fusion}$ denotes the output features of the proposed attentive fusion module. 
Note that the attentive fusion module is a part of the recurrent SR branch, so that the gradients can be back-propagated to both the SR and alignment branches in a recursive manner. Moreover, the landmark estimation can be supervised by not only the loss imposed on the recurrent alignment branch, but also by the revision of FSR results through the attentive fusion module.

\subsection{Objective Functions}

\textbf{Adversarial Loss}: Recently GAN~\cite{ledig2017photo,wang2018esrgan,chen2018fsrnet} has been successful in generative tasks, and is proven effective in recovering high-fidelity images. Hence we introduce the adversarial loss~\cite{ledig2017photo} to generate photo-realistic face images. We build a discriminator $D$ to differentiate the ground-truth and the super-resolved counterparts by minimizing
\begin{eqnarray}
\mathcal{L}^{Dis} =  -\mathbb{E}[\mathrm{log}(D(I^{HR}))]-\mathbb{E}[\mathrm{log}(1-D(G(I^{LR})))]. 
\end{eqnarray}
Meanwhile, the generator tries to fool the discriminator and minimizes
\begin{eqnarray}
\mathcal{L}^{Adv} &=&  -\mathbb{E}[\mathrm{log}(D(G(I^{LR})))]. 
\end{eqnarray}

\textbf{Perceptual Loss}: We also apply a perceptual loss to enhance the perceptual quality of SR images, similar to~\cite{ledig2017photo,chen2018fsrnet}. We employ a pretrained face recognition model, LightCNN~\cite{wu2018light} to extract features for images. The loss improves the perceptual similarity by reducing the euclidean distances between the features of SR and HR images, $\phi(I^{SR})$ and $\phi(I^{HR})$. Hence we define the perceptual loss as: 
\begin{eqnarray}
\mathcal{L}^{Perc} &=&  \mathbb{E}\|\phi(I^{SR})-\phi(I^{HR})\|_1. 
\end{eqnarray}

\textbf{Overall Objective}: The generator is optimized by minimizing the following overall objective function:
\begin{eqnarray}
\mathcal{L}^{G} &=& \mathcal{L}^{Pixel}+\lambda^{Adv}\cdot \mathcal{L}^{Adv} \\ \nonumber
&&+\lambda^{Perc}\cdot \mathcal{L}^{Perc}+\beta^{Align}\cdot \mathcal{L}^{Align},
\end{eqnarray}
where $\lambda^{Adv}$ and $\lambda^{Perc}$ denote the trade-off parameters for the adversarial loss and the perceptual loss, respectively. Since the recurrent alignment module is optimized as a part of the whole framework, the overall objective also includes this term of loss weighted by $\beta^{Align}$. 
For the training of our PSNR-oriented model DIC, we set $\lambda^{Adv}=\lambda^{Perc}=0$. Then complete losses are used to obtain the perceptual-pleasing model DICGAN.

\section{Experiments}

\subsection{Datasets and Metrics}

\begin{table}[tbp]
\centering
\begin{center}
   \caption{Comparison of PSNR and SSIM performance with state-of-the-art FSR methods. The best and second best performance is \textbf{highlighted} in \textcolor{red}{\textbf{red}} and \textcolor{blue}{\textbf{blue}}, respectively. }
   \vspace{5px}
   \label{tab:all_results_PSNR}
   \begin{tabular}{|c|c|c|c|c|}
   \hlineB{2.5}
   &\multicolumn{2}{c|}{CelebA}&\multicolumn{2}{c|}{Helen} \\
   \cline{2-5}
   Method&PSNR&SSIM&PSNR&SSIM \\
   \hline
   \hline
   Bicubic & 23.58 & 0.6285 & 23.89 & 0.6751 \\
   \hline
   SRResNet~\cite{ledig2017photo} & 25.82 & 0.7369 & 25.30 & 0.7297 \\
   \hline
   URDGN~\cite{yu2016ultra} & 24.63 & 0.6851 & 24.22 & 0.6909 \\
   \hline
   RDN~\cite{zhang2018residual} & 26.13 & 0.7412 & 25.34 & 0.7249 \\
   \hline
   PFSR~\cite{kim2019progressive} & 24.43 & 0.6991 & 24.73 & 0.7323 \\
   \hline
   FSRNet~\cite{chen2018fsrnet} & \textcolor{blue}{\textbf{26.48}} & \textcolor{blue}{\textbf{0.7718}} & 25.90 & \textcolor{blue}{\textbf{0.7759}} \\
   \hline
   FSRGAN~\cite{chen2018fsrnet} & 25.06 & 0.7311 & 24.99 & 0.7424 \\
   \hline
   DIC & \textcolor{red}{\textbf{27.37}} & \textcolor{red}{\textbf{0.7962}} & \textcolor{red}{\textbf{26.69}} & \textcolor{red}{\textbf{0.7933}} \\
   \hline
   DICGAN & 26.34 & 0.7562 & \textcolor{blue}{\textbf{25.96}} & 0.7624 \\
   \hlineB{2.5}
   \end{tabular}
\end{center}
\vspace{-5mm}
\end{table}

\begin{figure*}
\newlength\fsdurthree
\setlength{\fsdurthree}{-4mm}
\centering
\begin{adjustbox}{valign=t}
  \begin{tabular}{cccccccc}
Bicubic\hspace*{\fsdurthree} & RDN~\cite{zhang2018residual}\hspace*{\fsdurthree} & FSRNet~\cite{chen2018fsrnet}\hspace*{\fsdurthree} & FSRGAN~\cite{chen2018fsrnet}\hspace*{\fsdurthree} & PFSR~\cite{kim2019progressive}\hspace*{\fsdurthree} & DIC\hspace*{\fsdurthree} & DICGAN\hspace*{\fsdurthree} & HR\hspace*{\fsdurthree} \\
  \includegraphics[width=0.123\textwidth]{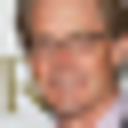}\hspace*{\fsdurthree} &
  \includegraphics[width=0.123\textwidth]{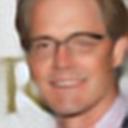}\hspace*{\fsdurthree} &
  \includegraphics[width=0.123\textwidth]{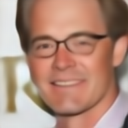}\hspace*{\fsdurthree} &
  \includegraphics[width=0.123\textwidth]{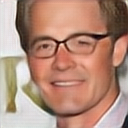}\hspace*{\fsdurthree} &
  \includegraphics[width=0.123\textwidth]{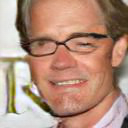}\hspace*{\fsdurthree} &
  \includegraphics[width=0.123\textwidth]{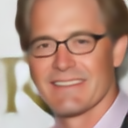}\hspace*{\fsdurthree} &
  \includegraphics[width=0.123\textwidth]{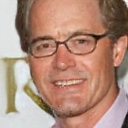}\hspace*{\fsdurthree} &
  \includegraphics[width=0.123\textwidth]{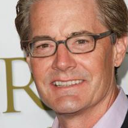}\hspace*{\fsdurthree} \\

  \includegraphics[width=0.123\textwidth]{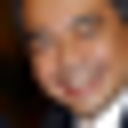}\hspace*{\fsdurthree} &
  \includegraphics[width=0.123\textwidth]{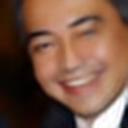}\hspace*{\fsdurthree} &
  \includegraphics[width=0.123\textwidth]{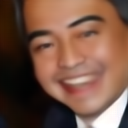}\hspace*{\fsdurthree} &
  \includegraphics[width=0.123\textwidth]{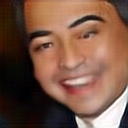}\hspace*{\fsdurthree} &
  \includegraphics[width=0.123\textwidth]{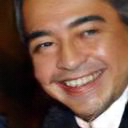}\hspace*{\fsdurthree} &
  \includegraphics[width=0.123\textwidth]{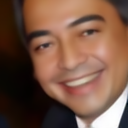}\hspace*{\fsdurthree} &
  \includegraphics[width=0.123\textwidth]{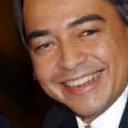}\hspace*{\fsdurthree} &
  \includegraphics[width=0.123\textwidth]{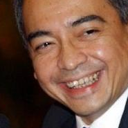}\hspace*{\fsdurthree} \\

  \includegraphics[width=0.123\textwidth]{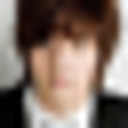}\hspace*{\fsdurthree} &
  \includegraphics[width=0.123\textwidth]{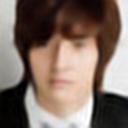}\hspace*{\fsdurthree} &
  \includegraphics[width=0.123\textwidth]{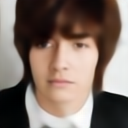}\hspace*{\fsdurthree} &
  \includegraphics[width=0.123\textwidth]{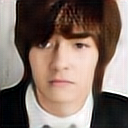}\hspace*{\fsdurthree} &
  \includegraphics[width=0.123\textwidth]{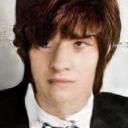}\hspace*{\fsdurthree} &
  \includegraphics[width=0.123\textwidth]{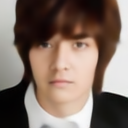}\hspace*{\fsdurthree} &
  \includegraphics[width=0.123\textwidth]{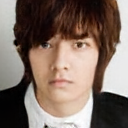}\hspace*{\fsdurthree} &
  \includegraphics[width=0.123\textwidth]{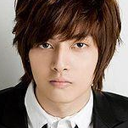}\hspace*{\fsdurthree} \\

  \includegraphics[width=0.123\textwidth]{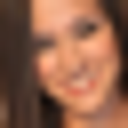}\hspace*{\fsdurthree} &
  \includegraphics[width=0.123\textwidth]{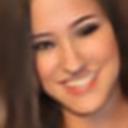}\hspace*{\fsdurthree} &
  \includegraphics[width=0.123\textwidth]{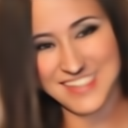}\hspace*{\fsdurthree} &
  \includegraphics[width=0.123\textwidth]{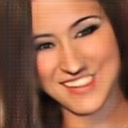}\hspace*{\fsdurthree} &
  \includegraphics[width=0.123\textwidth]{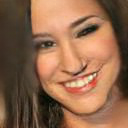}\hspace*{\fsdurthree} &
  \includegraphics[width=0.123\textwidth]{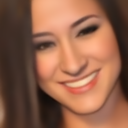}\hspace*{\fsdurthree} &
  \includegraphics[width=0.123\textwidth]{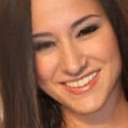}\hspace*{\fsdurthree} &
  \includegraphics[width=0.123\textwidth]{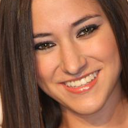}\hspace*{\fsdurthree} \\

  \includegraphics[width=0.123\textwidth]{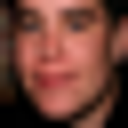}\hspace*{\fsdurthree} &
  \includegraphics[width=0.123\textwidth]{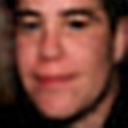}\hspace*{\fsdurthree} &
  \includegraphics[width=0.123\textwidth]{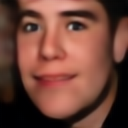}\hspace*{\fsdurthree} &
  \includegraphics[width=0.123\textwidth]{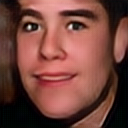}\hspace*{\fsdurthree} &
  \includegraphics[width=0.123\textwidth]{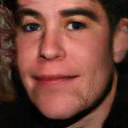}\hspace*{\fsdurthree} &
  \includegraphics[width=0.123\textwidth]{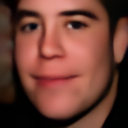}\hspace*{\fsdurthree} &
  \includegraphics[width=0.123\textwidth]{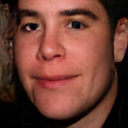}\hspace*{\fsdurthree} &
  \includegraphics[width=0.123\textwidth]{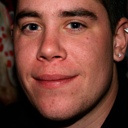}\hspace*{\fsdurthree} \\

  \includegraphics[width=0.123\textwidth]{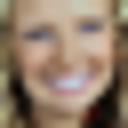}\hspace*{\fsdurthree} &
  \includegraphics[width=0.123\textwidth]{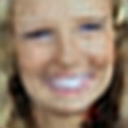}\hspace*{\fsdurthree} &
  \includegraphics[width=0.123\textwidth]{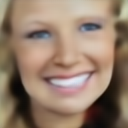}\hspace*{\fsdurthree} &
  \includegraphics[width=0.123\textwidth]{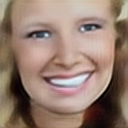}\hspace*{\fsdurthree} &
  \includegraphics[width=0.123\textwidth]{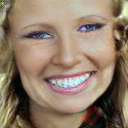}\hspace*{\fsdurthree} &
  \includegraphics[width=0.123\textwidth]{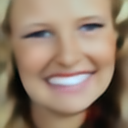}\hspace*{\fsdurthree} &
  \includegraphics[width=0.123\textwidth]{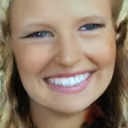}\hspace*{\fsdurthree} &
  \includegraphics[width=0.123\textwidth]{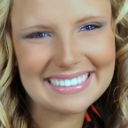}\hspace*{\fsdurthree} \\

  \end{tabular}
\end{adjustbox}
\vspace{1mm}
\caption{
	Visual comparison with state-of-the-art FSR methods. Other FSR methods may either produce structural distortions on key facial parts or present undesirable artifacts. Our proposed DIC and DICGAN methods have a significant advantage in handling large pose and rotation variations. The qualitative comparison indicates the proposed method outperforms other FSR methods. 
	Best viewed on screen.  
}
\label{fig:all_visual}
\vspace{-2mm}
\end{figure*}

We conduct experiments on two widely used face datasets: CelebA~\cite{liu2015deep} and Helen~\cite{le2012interactive}. For both datasets we use OpenFace~\cite{baltrusaitis2018openface, zadeh2017convolutional, baltrusaitis2013constrained} to detect 68 landmarks as ground-truth. Based on the estimated landmarks, we crop square regions in each image to remove the background and resize them to 128$\times$128 pixels without any pre-alignment. Then we downsample these HR images into 16$\times$16 LR inputs with bicubic degradation. For CelebA dataset, we use 168854 images for training and 1000 images for testing. For Helen dataset, we use 2005 images for training and 50 images for testing.

SR results are evaluated with PSNR and SSIM~\cite{wang2004image}. They are computed on the Y channel of transformed YCbCr space. We also use face alignment as a metric to measure the accuracy of face recovery. We use a pretrained HourGlass network to detect the face landmarks and use Normalized Root Mean Squared Error (NRMSE) to evaluate landmark estimation results. In our experiment, NRMSE is normalized by the width of the face.

\subsection{Implementation Details}

\textbf{Training Setting} The architecture of the recurrent SR module follows the feedback block in \cite{li2019feedback}. We set the number of groups to 6, the number of steps to 4 and the number of feature channels to 48. For Helen, data augmentation is performed on training images, which are randomly rotated by $90^\circ$, $180^\circ$, $270^\circ$ and flipped horizontally.  We train the PSNR-oriented model with the pixel loss and the alignment loss weighted by $\beta^{Align}=0.1$. For GAN training, we use the pretrained PSNR-oriented parameters as initialization and train the model with $\lambda^{Adv}=0.005$ and $\lambda^{Perc}=0.1$. The model is trained by ADAM optimizer~\cite{kingma2014adam} with $\beta_1=0.9, \beta_2=0.999$ and $\epsilon=10^{-8}$. The initial learning rate is $1\times10^{-4}$ and is halved at $1\times10^{4}, 2\times10^{4}, 4\times10^{4}$ iterations. Our experiments are implemented on Pytorch~\cite{paszke2017automatic} with NVIDIA RTX 2080Ti GPUs.

\subsection{Results and Analysis}

\textbf{Comparison with the State-of-the-Arts}: We compare our proposed DIC method with state-of-the-art FSR methods.
Table~\ref{tab:all_results_PSNR} tabulates the quantitative results on CelebA and Helen. It can be observed that our DIC method achieves the best PSNR and SSIM performance on both datasets. It is noteworthy that DIC outperforms FSRNet by a large margin. Therefore, our method obtains better inference by the progressive collaboration between the SR and alignment processes. 
Moreover, DICGAN gets comparable performance with FSRNet which is a PSNR-oriented method. This indicates that our DICGAN method is able to preserve pixel-wise accuracy while increasing perceptual quality of the super-resolved images. 

We visualize some SR results of different methods as shown in Figure~\ref{fig:all_visual}. We see that DIC recovers correct details while other methods fail in giving pleasant results. This indicates that our method is able to produce more stable SR results than other methods. Note that our method has a significant advantage in handling large pose and rotation variations. 
The reason is that the iterative alignment block can predict progressively more accurate landmarks to guide the reconstruction in each step. Therefore our method performs better in preserving facial structures and generating better details even though faces have large pose and rotation.  Furthermore, DICGAN produces more realistic textures of images while other methods yield severe artifacts and distortions. Therefore, the qualitative comparison with state-of-the-art face SR methods demonstrates the powerful generative ability of our methods.

\begin{figure}[t]
\setlength{\fsdurthree}{-3mm}
\centering
   \begin{adjustbox}{valign=t}
   \tiny
      \begin{tabular}{ccc}
         
         \includegraphics[width=0.14\textwidth]{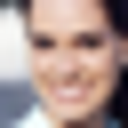}& 
         \includegraphics[width=0.14\textwidth]{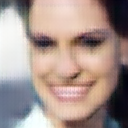}\hspace*{\fsdurthree}& 
         \includegraphics[width=0.14\textwidth]{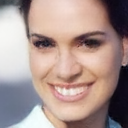}\hspace*{\fsdurthree}\\
         Bicubic& Step 1\hspace*{\fsdurthree}&Step 2\hspace*{\fsdurthree}\\
         PSNR/SSIM& 24.73/0.7605\hspace*{\fsdurthree} & 27.94/0.8667\hspace*{\fsdurthree} \\
         NRMSE& 0.0265\hspace*{\fsdurthree} & 0.0211\hspace*{\fsdurthree} \\
         \includegraphics[width=0.14\textwidth]{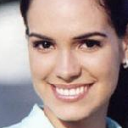}&
         \includegraphics[width=0.14\textwidth]{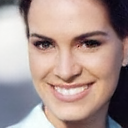}\hspace*{\fsdurthree}&
         \includegraphics[width=0.14\textwidth]{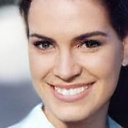}\hspace*{\fsdurthree}\\
         HR&Step 3\hspace*{\fsdurthree} &Step 4\hspace*{\fsdurthree}\\
         PSNR/SSIM& 28.32/0.8782\hspace*{\fsdurthree} & 28.32/0.8791\hspace*{\fsdurthree} \\
         NRMSE& 0.0204\hspace*{\fsdurthree} & 0.0194\hspace*{\fsdurthree} \\
         \vspace{1.5mm}\\

         \includegraphics[width=0.14\textwidth]{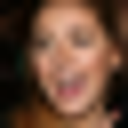}& 
         \includegraphics[width=0.14\textwidth]{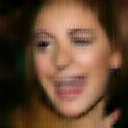}\hspace*{\fsdurthree}& 
         \includegraphics[width=0.14\textwidth]{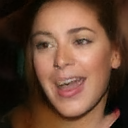}\hspace*{\fsdurthree}\\
         Bicubic& Step 1\hspace*{\fsdurthree}&Step 2\hspace*{\fsdurthree}\\
         PSNR/SSIM& 26.35/0.7690\hspace*{\fsdurthree} & 27.38/0.8212\hspace*{\fsdurthree} \\
         NRMSE& 0.0293\hspace*{\fsdurthree} & 0.0266\hspace*{\fsdurthree} \\
         \includegraphics[width=0.14\textwidth]{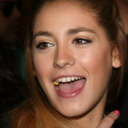}&
         \includegraphics[width=0.14\textwidth]{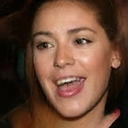}\hspace*{\fsdurthree}&
         \includegraphics[width=0.14\textwidth]{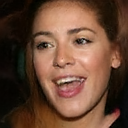}\hspace*{\fsdurthree}\\
         HR&Step 3\hspace*{\fsdurthree} &Step 4\hspace*{\fsdurthree}\\
         PSNR/SSIM& 27.90/0.8316\hspace*{\fsdurthree} & 28.07/0.8350\hspace*{\fsdurthree} \\
         NRMSE& 0.0260\hspace*{\fsdurthree} & 0.0249\hspace*{\fsdurthree} \\
      \end{tabular}
   \end{adjustbox}
\vspace{2mm}
\caption{
		Visual comparison of different steps. With the iterative collaboration, visual quality and quantitative measurement both get better progressively.   
	}
\label{fig:step_visual}
\vspace{-5mm}
\end{figure}

Similar to~\cite{chen2018fsrnet}, we conduct face alignment as a measurement to evaluate the quality of the super-resolved images. We adopt a pretrained face alignment model with four stacked hourglass modules~\cite{newell2016stacked}. The alignment accuracy is reflected by a widely used metric NRMSE. Lower NRMSE values reveal better alignment accuracy and higher quality of SR images. Table~\ref{tab:results_landmark} shows the NRMSE values of our methods and other compared SR methods. We can see our DICGAN method outperforms other methods on both datasets. While other SR methods also use facial priors such as landmarks and component maps, the prior information is estimated from the input LR face images or coarsely recovered ones where facial structures are severely unclear and degraded. Hence such facial priors can provide limited guidance to the reconstruction procedure. Consequently, recovered images may also contain corresponding structural incorrectness. Differently, our method revises the landmark estimation in every step for providing more accurate auxiliary information to the SR branch. Meanwhile, the attentive fusion module can integrate the prior guidance effectively to boost the final performance.  

\textbf{User Study}: We also conduct a user study as a subjective assessment to further evaluate our SR quality compared to previous face SR methods. Details are described in the supplementary material. 

\begin{table}[tbp]
\centering
\begin{center}
   \caption{Comparison of NRMSE performance with state-of-the-art FSR methods.
   The best and second best performance is \textbf{highlighted} in \textcolor{red}{\textbf{red}} and \textcolor{blue}{\textbf{blue}}, respectively}
   \vspace{5px}
   \label{tab:results_landmark}
   \begin{tabular}{|c|c|c|}
   \hlineB{2.5}
   Method&CelebA&Helen\\
   \hline
   \hline
   Bicubic & 0.3385 & 0.4577 \\
   \hline
   RDN~\cite{zhang2018residual} & 0.1415 & 0.4437 \\
   \hline
   PFSR~\cite{kim2019progressive} & 0.1917 & 0.3498 \\
   \hline
   FSRNet~\cite{chen2018fsrnet} & 0.1430 & 0.3723 \\
   \hline
   FSRGAN~\cite{chen2018fsrnet} & 0.1463 & \textcolor{blue}{\textbf{0.3408}} \\
   \hline
   DIC & \textcolor{blue}{\textbf{0.1320}} & 0.3674 \\
   \hline
   DICGAN & \textcolor{red}{\textbf{0.1319}} & \textcolor{red}{\textbf{0.3336}} \\
   \hlineB{2.5}
   \end{tabular}
\end{center}
\vspace{-5mm}
\end{table}

\begin{table}[tbp]
\centering
\begin{center}
   \caption{Quantitative comparison of different steps on CelebA. The best results are \textbf{highlighted}. }
   \vspace{5px}
   \label{tab:results_step_CelebA}
   \begin{tabular}{|c|c|c|c|c|}
   \hlineB{2.5}
   Metric&Step 1&Step 2&Step 3&Step 4\\
   \hline
   \hline
   PSNR & 24.41 & 25.71 & 26.30 & \textbf{26.34} \\
   \hline
   SSIM & 0.6688 & 0.7180 & 0.7521 &\textbf{0.7561}\\
   \hline
   NRMSE & 0.0322 & 0.0306 & 0.0285 & \textbf{0.0273} \\
   \hlineB{2.5}
   \end{tabular}
\end{center}
\vspace{-5mm}
\end{table}

\begin{table}[tbp]
\centering
\begin{center}
   \caption{Quantitative comparison of different steps on Helen. The best results are \textbf{highlighted}. }
   \vspace{5px}
   \label{tab:results_step_Helen}
   \begin{tabular}{|c|c|c|c|c|}
   \hlineB{2.5}
   Metric&Step 1&Step 2&Step 3&Step 4\\
   \hline
   \hline
   PSNR & 24.88 & 25.45 & 25.96 & \textbf{25.96} \\
   \hline
   SSIM & 0.7094 & 0.7332 & 0.7587 & \textbf{0.7624}\\
   \hline
   NRMSE & 0.1057 & 0.0854 & 0.0837 & \textbf{0.0520} \\
   \hlineB{2.5}
   \end{tabular}
\end{center}
\vspace{-5mm}
\end{table}

\textbf{Study of Iterative Learning}: To better show the merits of the proposed scheme of iterative collaboration, we also evaluate the quality of the SR outputs. As mentioned above, we use PSNR, SSIM and NRMSE as measurement metrics. Differently, in this experiment, NRMSE is computed by the landmarks estimated by the alignment branch in the corresponding steps. The performance on CelebA and Helen is presented in Table~\ref{tab:results_step_CelebA} and Table~\ref{tab:results_step_Helen}, respectively. We can see from step 1 to step 4, the performance gets better progressively. It is noteworthy that the NRMSE values in Table~\ref{tab:results_step_CelebA} and Table~\ref{tab:results_step_Helen} are much lower than those in Table~\ref{tab:results_landmark}. In fact, in our alignment branch, the parameters are much fewer than the stacked hourglass model which is used to estimate landmarks in Table~\ref{tab:results_landmark}. The reason why our model gets more accurate alignment results with fewer parameters is that our model can learn to capture face structures in different-level super-resolved images. Due to this ability, our model can provide relatively accurate landmarks in each step for better collaboration. Therefore, the comparison proves that our method is able to achieve progressively better SR quality and landmark estimation simultaneously.

Furthermore, visual comparison of different steps are shown in Figure~\ref{fig:step_visual}. The results show the generation of facial components are improved step by step. In the last step, our model obtains geometric-pleasing and high-fidelity SR images. From the PSNR, SSIM and NRMSE values in each step, we can also see the consistent improvement of our scheme of iterative collaboration. Moreover, from Table~\ref{tab:results_step_CelebA}, Table~\ref{tab:results_step_Helen} and Figure~\ref{fig:step_visual}, three steps may be a suitable choice for good enough recovery and efficient computation. 

\begin{figure}
\setlength{\fsdurthree}{-3mm}
\centering
\begin{adjustbox}{valign=t}
   \tiny
   \begin{tabular}{ccccc}
   \includegraphics[width=0.085\textwidth]{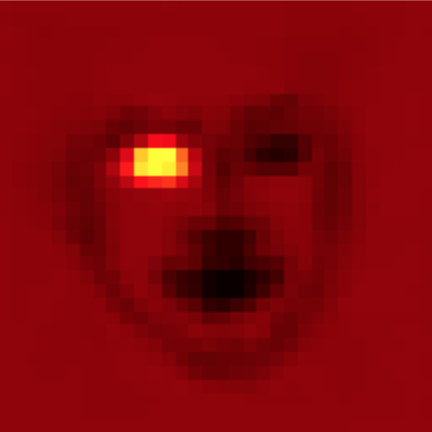}\hspace*{\fsdurthree} &
   \includegraphics[width=0.085\textwidth]{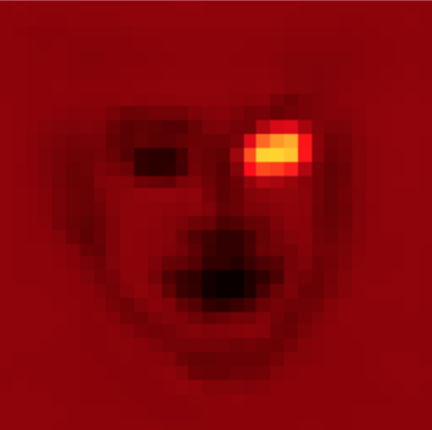}\hspace*{\fsdurthree} &
   \includegraphics[width=0.085\textwidth]{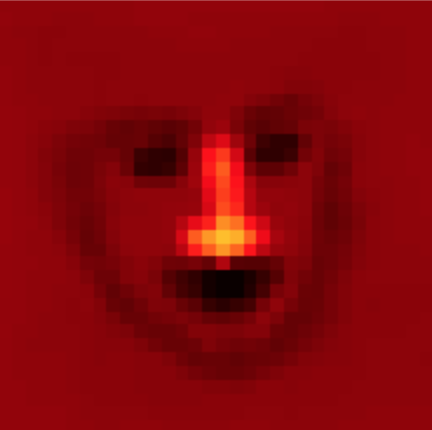}\hspace*{\fsdurthree} &
   \includegraphics[width=0.085\textwidth]{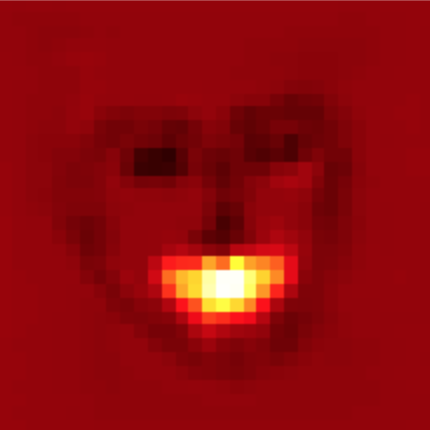}\hspace*{\fsdurthree} &
   \includegraphics[width=0.085\textwidth]{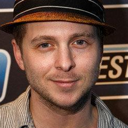}\hspace*{\fsdurthree}\\
   left eye attention\hspace*{\fsdurthree} & right eye attention\hspace*{\fsdurthree} & nose attention\hspace*{\fsdurthree} & mouth attention\hspace*{\fsdurthree} & HR\hspace*{\fsdurthree} \\
   
   \includegraphics[width=0.085\textwidth]{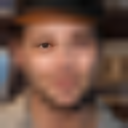}\hspace*{\fsdurthree} &
   \includegraphics[width=0.085\textwidth]{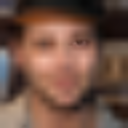}\hspace*{\fsdurthree} &
   \includegraphics[width=0.085\textwidth]{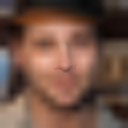}\hspace*{\fsdurthree} &
   \includegraphics[width=0.085\textwidth]{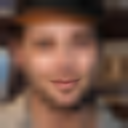}\hspace*{\fsdurthree} &
   \includegraphics[width=0.085\textwidth]{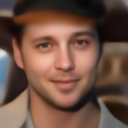}\hspace*{\fsdurthree} \\
   left eye image\hspace*{\fsdurthree} & right eye image\hspace*{\fsdurthree} & nose image\hspace*{\fsdurthree} & mouth image\hspace*{\fsdurthree} & SR\hspace*{\fsdurthree} \\
   
   \end{tabular}
\end{adjustbox}
\vspace{1mm}
\caption{
		Visual effects of the proposed attentive fusion module. The first row displays the attention maps and the ground-truth image.  The second row presents the SR outputs recovered by the features of the corresponding facial components. The component-specialized generation demonstrates the effectiveness of the proposed attentive fusion module. 
	}
\label{fig:attention_visual}
\end{figure}

\textbf{Effects of Attentive Fusion}: We implement another experiment to better investigate the effectiveness of our proposed attentive fusion module. Since we use group convolution layers to extract specialized representations for different facial parts, we only remain the representation of one part and visualize the SR results as shown in Figure~\ref{fig:attention_visual}. For a certain component, we remove features of the other components by setting the corresponding attention maps to 0. By this means, the final outputs only contain accurate information for one facial component. From Figure~\ref{fig:attention_visual}, we can indeed see different parts can be recovered separately by the representations. The results demonstrate the advantages of the proposed attentive fusion module, which can explicitly guide the component-specialized generation in an efficient and flexible way. 

\textbf{Ablation Study}: We further implement an ablation study to measure the effectiveness of the iterative collaboration framework and the attentive fusion module. On the one hand, in order to validate the effects of facial priors, we remove the alignment branch and the attentive fusion module. This model is called DIC-NL, which is equivalent to a recurrent network for single image super-resolution without the prior information of landmark maps. On the other hand, we remove the attentive fusion module and concatenate landmarks (CL) to evaluate the effects of the proposed fusion module quantitatively. This model is denoted as DIC-CL. PSNR and SSIM performance on the dataset of CelebA is presented in Table~\ref{tab:ablation}. From the table we can see when the SR network loses the guidance provided by face landmarks, SR quality is degraded severely since its ability to capture facial structural configuration is weakened. Moreover, DIC-CL has an advantage over DIC-NL since it incorporates the prior information by concatenation. A large enhancement can also be observed due to the integration. However, the SR performance of DIC-CL is still far from that of the DIC method. The reason is that concatenating landmark maps is an implicit knowledge to face SR and is limited in providing adequate guidance. Differently, our DIC method not only integrates the structural knowledge, but also explicitly induces the component-specialized feature extraction for more photo-realistic SR images. Hence the results prove the superiority of the proposed method.

\begin{table}[tbp]
\centering
\begin{center}
   \caption{Quantitative comparison of different models. The best results are \textbf{highlighted}. (NL: no landmarks, CL: concatenated landmarks.) }
   \vspace{5px}
   \label{tab:ablation}
   \begin{tabular}{|c|c|c|}
   \hlineB{2.5}
   Method&PSNR&SSIM \\
   \hline
   \hline
   DIC-NL & 26.31 & 0.7526 \\
   \hline
   DIC-CL & 26.93 & 0.7811 \\
   \hline
   DIC & \textbf{27.37} & \textbf{0.7962} \\
   \hlineB{2.5}
   \end{tabular}
\end{center}
\vspace{-5mm}
\end{table}

\section{Conclusion}

In this paper, we have proposed a deep iterative collaboration network for face super-resolution. Specifically, a recurrent SR branch collaborates with a recurrent alignment branch to recover high-quality face SR images iteratively and progressively. In each step, the SR process utilizes the estimated landmarks from the alignment branch to produce better face images which are important for the alignment branch to estimate more accurate landmarks.
Furthermore, we have proposed a new attentive fusion module to exploit attention maps and extract individual features for each facial component according to the estimated landmarks. Quantitative and qualitative results of face SR on two widely-used benchmark datasets have demonstrated the effectiveness of the proposed method. 

\section*{Acknowledgement}
This work was supported in part by the National Key Research and Development Program of China under Grant 2017YFA0700802, in part by the National Natural Science Foundation of China under Grant 61822603, Grant U1813218, Grant U1713214, and Grant 61672306, in part by the Shenzhen Fundamental Research Fund (Subject Arrangement) under Grant JCYJ20170412170602564, and in part by Tsinghua University Initiative Scientfic Research Program.

{\small
\bibliographystyle{ieee_fullname}
\bibliography{egbib}
}

\clearpage

\appendix

\section*{Supplementary Material}

\section{More Details on Network Architecture}
\label{sec:A}

Here we describe more details of our recurrent networks. Table~\ref{tab:SR_branch} shows the detailed architecture of the SR branch. Given input LR images, LR features are extracted by $G_1$ and are subsequently concatenated with the feedback features. Then through $G_R$, which consists of a convolutional layer, an attentive fusion module and a recurrent SR module, the obtained features are used as both the feedback signals and the features for the following generation. Finally, SR images are recovered by the generation layers $G_2$ and the addition operation. $G_2$ is comprised of a deconvolutional layer with a kernel size of 8 and a convolutional layer. 

Besides, Table~\ref{tab:Align_branch} presents the details of our recurrent alignment branch. $A_1$ and $A_2$ are the pre-processing and post-processing blocks, which have the same architecture as those in~\cite{newell2016stacked} except that the batch normalization layers are removed. The recurrent hourglass module has similar architecture to the single hourglass module in~\cite{newell2016stacked}. Differently, the input and output of $A_R$ both include two components. The input is obtained by concatenating the pre-processing feature with the feedback feature while the output is split into two parts, a feedback feature and a feature for the final landmark estimation.

\section{User Study}
\label{sec:B}

\begin{figure}
    \centering
    \includegraphics[width=\linewidth]{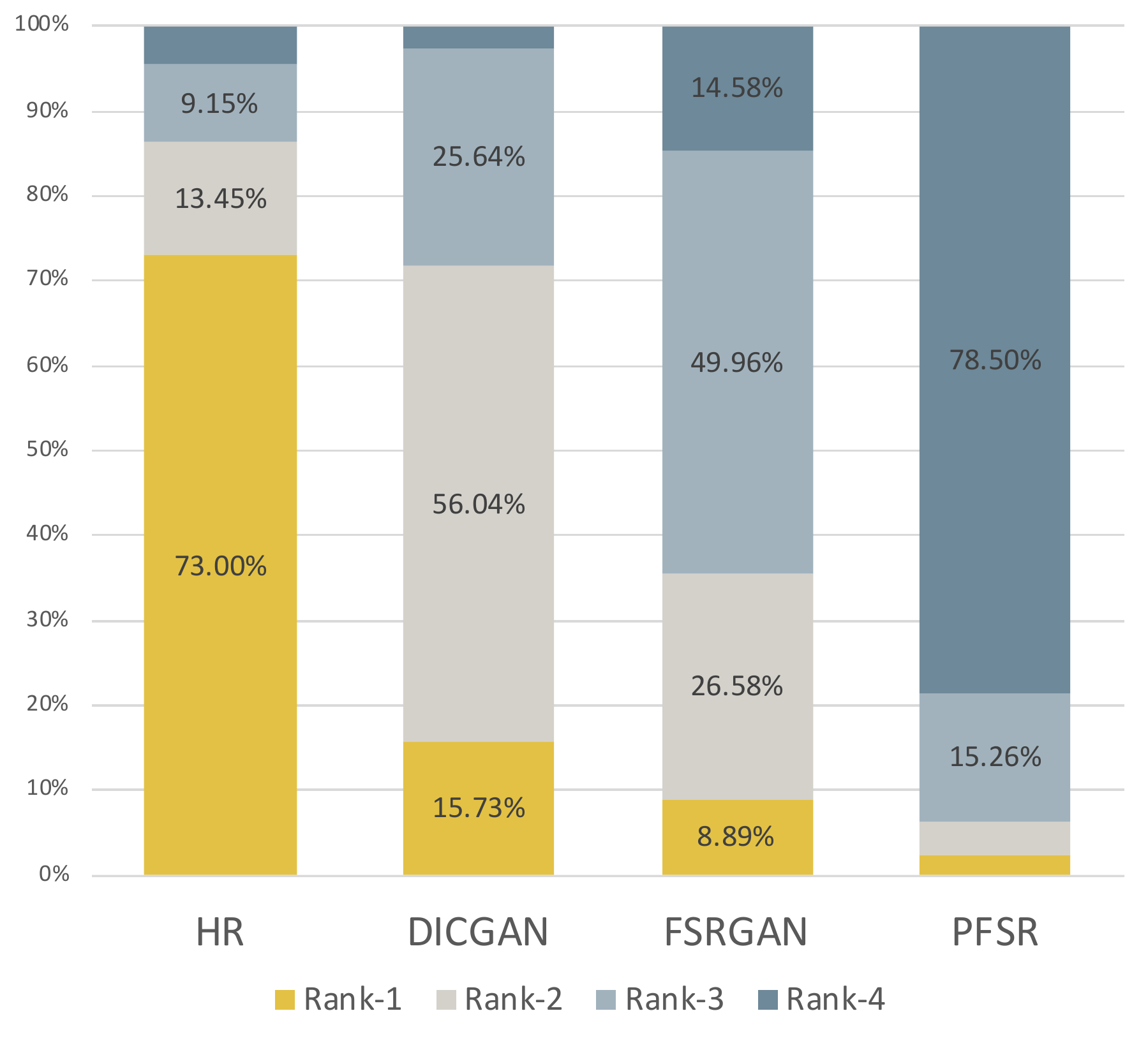}
    \caption{Results of the user study. Our method performs better than state-of-the-art FSR methods in recovering perceptual-pleasant face images. }
    \label{fig:user-study}
\end{figure}

We conduct a user study to further evaluate the visual quality of the super-resolved images. We randomly select 30 images from the testing set of CelebA~\cite{liu2015deep} and display the corresponding SR results of our DICGAN, FSRGAN~\cite{chen2018fsrnet}, PFSR~\cite{kim2019progressive} and the HR images in a random order. 39 human raters are asked to rank these four versions of images in terms of perceptual satisfaction. The results are shown in Figure~\ref{fig:user-study}. As expected, most of the HR images are regarded as the best among the four versions. Moreover, our DICGAN obtains much more votes of rank-1 and rank-2 than FSRGAN and PFSR, which means the proposed method outperforms the state-of-the-art face SR methods by a large margin. 
We observe that PFSR scores the worst among three FSR methods. 
We think the reason is that PFSR mainly focuses on well-aligned face images. Hence when the input faces are with large variations of pose and rotation, PFSR fails to present satisfactory SR results.

\section{Visual Results}

\begin{figure*}
\setlength{\fsdurthree}{-4mm}
\centering
\begin{adjustbox}{valign=t}
  \begin{tabular}{cccccccc}

\multicolumn{8}{c}{201448 from CelebA} \\
\includegraphics[width=0.123\textwidth]{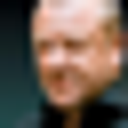}\hspace*{\fsdurthree} &
\includegraphics[width=0.123\textwidth]{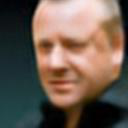}\hspace*{\fsdurthree} &
\includegraphics[width=0.123\textwidth]{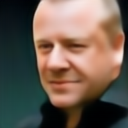}\hspace*{\fsdurthree} &
\includegraphics[width=0.123\textwidth]{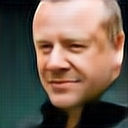}\hspace*{\fsdurthree} &
\includegraphics[width=0.123\textwidth]{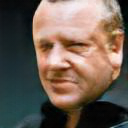}\hspace*{\fsdurthree} &
\includegraphics[width=0.123\textwidth]{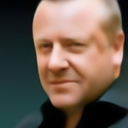}\hspace*{\fsdurthree} &
\includegraphics[width=0.123\textwidth]{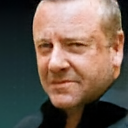}\hspace*{\fsdurthree} &
\includegraphics[width=0.123\textwidth]{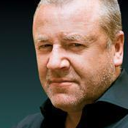}\hspace*{\fsdurthree} \\
Bicubic\hspace*{\fsdurthree} & RDN\hspace*{\fsdurthree} & FSRNet\hspace*{\fsdurthree} & FSRGAN\hspace*{\fsdurthree} & PFSR\hspace*{\fsdurthree} & DIC\hspace*{\fsdurthree} & DICGAN\hspace*{\fsdurthree} & HR\hspace*{\fsdurthree} \\
[4mm]
\multicolumn{8}{c}{201475 from CelebA} \\
\includegraphics[width=0.123\textwidth]{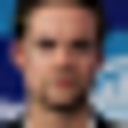}\hspace*{\fsdurthree} &
\includegraphics[width=0.123\textwidth]{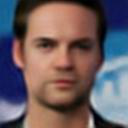}\hspace*{\fsdurthree} &
\includegraphics[width=0.123\textwidth]{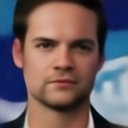}\hspace*{\fsdurthree} &
\includegraphics[width=0.123\textwidth]{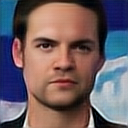}\hspace*{\fsdurthree} &
\includegraphics[width=0.123\textwidth]{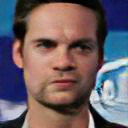}\hspace*{\fsdurthree} &
\includegraphics[width=0.123\textwidth]{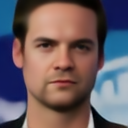}\hspace*{\fsdurthree} &
\includegraphics[width=0.123\textwidth]{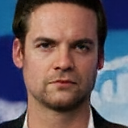}\hspace*{\fsdurthree} &
\includegraphics[width=0.123\textwidth]{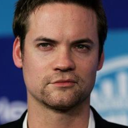}\hspace*{\fsdurthree} \\
Bicubic\hspace*{\fsdurthree} & RDN\hspace*{\fsdurthree} & FSRNet\hspace*{\fsdurthree} & FSRGAN\hspace*{\fsdurthree} & PFSR\hspace*{\fsdurthree} & DIC\hspace*{\fsdurthree} & DICGAN\hspace*{\fsdurthree} & HR\hspace*{\fsdurthree} \\
[4mm]
\multicolumn{8}{c}{201589 from CelebA} \\
\includegraphics[width=0.123\textwidth]{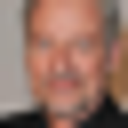}\hspace*{\fsdurthree} &
\includegraphics[width=0.123\textwidth]{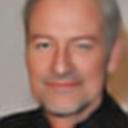}\hspace*{\fsdurthree} &
\includegraphics[width=0.123\textwidth]{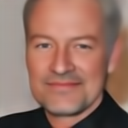}\hspace*{\fsdurthree} &
\includegraphics[width=0.123\textwidth]{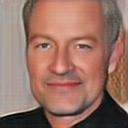}\hspace*{\fsdurthree} &
\includegraphics[width=0.123\textwidth]{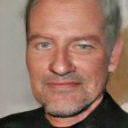}\hspace*{\fsdurthree} &
\includegraphics[width=0.123\textwidth]{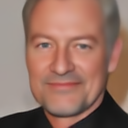}\hspace*{\fsdurthree} &
\includegraphics[width=0.123\textwidth]{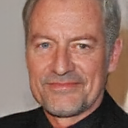}\hspace*{\fsdurthree} &
\includegraphics[width=0.123\textwidth]{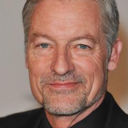}\hspace*{\fsdurthree} \\
Bicubic\hspace*{\fsdurthree} & RDN\hspace*{\fsdurthree} & FSRNet\hspace*{\fsdurthree} & FSRGAN\hspace*{\fsdurthree} & PFSR\hspace*{\fsdurthree} & DIC\hspace*{\fsdurthree} & DICGAN\hspace*{\fsdurthree} & HR\hspace*{\fsdurthree} \\
[4mm]
\multicolumn{8}{c}{202085 from CelebA} \\
\includegraphics[width=0.123\textwidth]{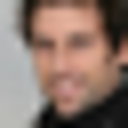}\hspace*{\fsdurthree} &
\includegraphics[width=0.123\textwidth]{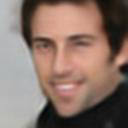}\hspace*{\fsdurthree} &
\includegraphics[width=0.123\textwidth]{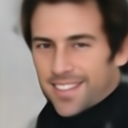}\hspace*{\fsdurthree} &
\includegraphics[width=0.123\textwidth]{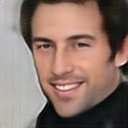}\hspace*{\fsdurthree} &
\includegraphics[width=0.123\textwidth]{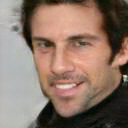}\hspace*{\fsdurthree} &
\includegraphics[width=0.123\textwidth]{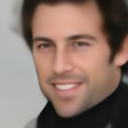}\hspace*{\fsdurthree} &
\includegraphics[width=0.123\textwidth]{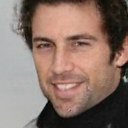}\hspace*{\fsdurthree} &
\includegraphics[width=0.123\textwidth]{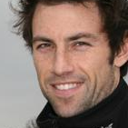}\hspace*{\fsdurthree} \\
Bicubic\hspace*{\fsdurthree} & RDN\hspace*{\fsdurthree} & FSRNet\hspace*{\fsdurthree} & FSRGAN\hspace*{\fsdurthree} & PFSR\hspace*{\fsdurthree} & DIC\hspace*{\fsdurthree} & DICGAN\hspace*{\fsdurthree} & HR\hspace*{\fsdurthree} \\
[4mm]
\multicolumn{8}{c}{202301 from CelebA} \\
\includegraphics[width=0.123\textwidth]{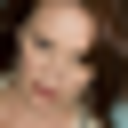}\hspace*{\fsdurthree} &
\includegraphics[width=0.123\textwidth]{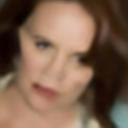}\hspace*{\fsdurthree} &
\includegraphics[width=0.123\textwidth]{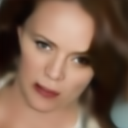}\hspace*{\fsdurthree} &
\includegraphics[width=0.123\textwidth]{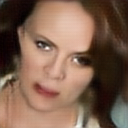}\hspace*{\fsdurthree} &
\includegraphics[width=0.123\textwidth]{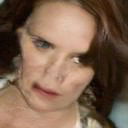}\hspace*{\fsdurthree} &
\includegraphics[width=0.123\textwidth]{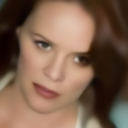}\hspace*{\fsdurthree} &
\includegraphics[width=0.123\textwidth]{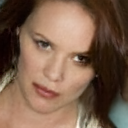}\hspace*{\fsdurthree} &
\includegraphics[width=0.123\textwidth]{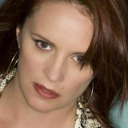}\hspace*{\fsdurthree} \\
Bicubic\hspace*{\fsdurthree} & RDN\hspace*{\fsdurthree} & FSRNet\hspace*{\fsdurthree} & FSRGAN\hspace*{\fsdurthree} & PFSR\hspace*{\fsdurthree} & DIC\hspace*{\fsdurthree} & DICGAN\hspace*{\fsdurthree} & HR\hspace*{\fsdurthree} \\
[4mm]
\multicolumn{8}{c}{201936 from CelebA} \\
\includegraphics[width=0.123\textwidth]{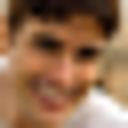}\hspace*{\fsdurthree} &
\includegraphics[width=0.123\textwidth]{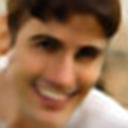}\hspace*{\fsdurthree} &
\includegraphics[width=0.123\textwidth]{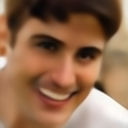}\hspace*{\fsdurthree} &
\includegraphics[width=0.123\textwidth]{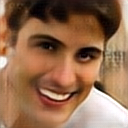}\hspace*{\fsdurthree} &
\includegraphics[width=0.123\textwidth]{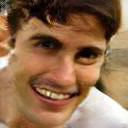}\hspace*{\fsdurthree} &
\includegraphics[width=0.123\textwidth]{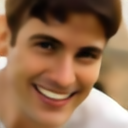}\hspace*{\fsdurthree} &
\includegraphics[width=0.123\textwidth]{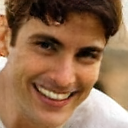}\hspace*{\fsdurthree} &
\includegraphics[width=0.123\textwidth]{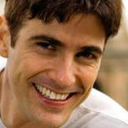}\hspace*{\fsdurthree} \\
Bicubic\hspace*{\fsdurthree} & RDN\hspace*{\fsdurthree} & FSRNet\hspace*{\fsdurthree} & FSRGAN\hspace*{\fsdurthree} & PFSR\hspace*{\fsdurthree} & DIC\hspace*{\fsdurthree} & DICGAN\hspace*{\fsdurthree} & HR\hspace*{\fsdurthree} \\

\end{tabular}
\end{adjustbox}
\vspace{1mm}

\caption{Qualitative comparison with state-of-the-art face super-resolution methods. }
\label{fig:supp_vis1}
\end{figure*}

\begin{figure*}
\setlength{\fsdurthree}{-4mm}
\centering
\begin{adjustbox}{valign=t}
\begin{tabular}{cccccccc}

\multicolumn{8}{c}{201937 from CelebA} \\
\includegraphics[width=0.123\textwidth]{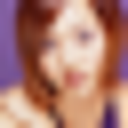}\hspace*{\fsdurthree} &
\includegraphics[width=0.123\textwidth]{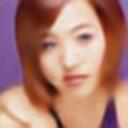}\hspace*{\fsdurthree} &
\includegraphics[width=0.123\textwidth]{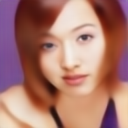}\hspace*{\fsdurthree} &
\includegraphics[width=0.123\textwidth]{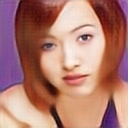}\hspace*{\fsdurthree} &
\includegraphics[width=0.123\textwidth]{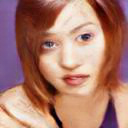}\hspace*{\fsdurthree} &
\includegraphics[width=0.123\textwidth]{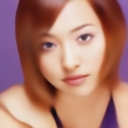}\hspace*{\fsdurthree} &
\includegraphics[width=0.123\textwidth]{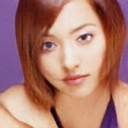}\hspace*{\fsdurthree} &
\includegraphics[width=0.123\textwidth]{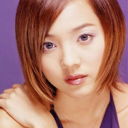}\hspace*{\fsdurthree} \\
Bicubic\hspace*{\fsdurthree} & RDN\hspace*{\fsdurthree} & FSRNet\hspace*{\fsdurthree} & FSRGAN\hspace*{\fsdurthree} & PFSR\hspace*{\fsdurthree} & DIC\hspace*{\fsdurthree} & DICGAN\hspace*{\fsdurthree} & HR\hspace*{\fsdurthree} \\
[4mm]
\multicolumn{8}{c}{201940 from CelebA} \\
\includegraphics[width=0.123\textwidth]{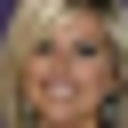}\hspace*{\fsdurthree} &
\includegraphics[width=0.123\textwidth]{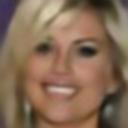}\hspace*{\fsdurthree} &
\includegraphics[width=0.123\textwidth]{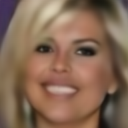}\hspace*{\fsdurthree} &
\includegraphics[width=0.123\textwidth]{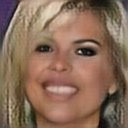}\hspace*{\fsdurthree} &
\includegraphics[width=0.123\textwidth]{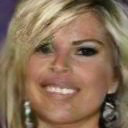}\hspace*{\fsdurthree} &
\includegraphics[width=0.123\textwidth]{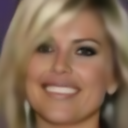}\hspace*{\fsdurthree} &
\includegraphics[width=0.123\textwidth]{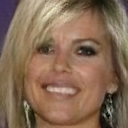}\hspace*{\fsdurthree} &
\includegraphics[width=0.123\textwidth]{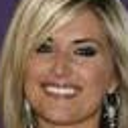}\hspace*{\fsdurthree} \\
Bicubic\hspace*{\fsdurthree} & RDN\hspace*{\fsdurthree} & FSRNet\hspace*{\fsdurthree} & FSRGAN\hspace*{\fsdurthree} & PFSR\hspace*{\fsdurthree} & DIC\hspace*{\fsdurthree} & DICGAN\hspace*{\fsdurthree} & HR\hspace*{\fsdurthree} \\
[4mm]
\multicolumn{8}{c}{201941 from CelebA} \\
\includegraphics[width=0.123\textwidth]{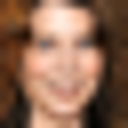}\hspace*{\fsdurthree} &
\includegraphics[width=0.123\textwidth]{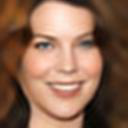}\hspace*{\fsdurthree} &
\includegraphics[width=0.123\textwidth]{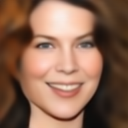}\hspace*{\fsdurthree} &
\includegraphics[width=0.123\textwidth]{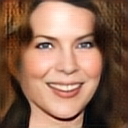}\hspace*{\fsdurthree} &
\includegraphics[width=0.123\textwidth]{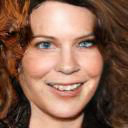}\hspace*{\fsdurthree} &
\includegraphics[width=0.123\textwidth]{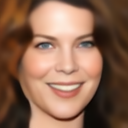}\hspace*{\fsdurthree} &
\includegraphics[width=0.123\textwidth]{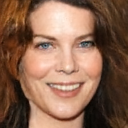}\hspace*{\fsdurthree} &
\includegraphics[width=0.123\textwidth]{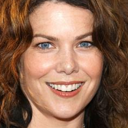}\hspace*{\fsdurthree} \\
Bicubic\hspace*{\fsdurthree} & RDN\hspace*{\fsdurthree} & FSRNet\hspace*{\fsdurthree} & FSRGAN\hspace*{\fsdurthree} & PFSR\hspace*{\fsdurthree} & DIC\hspace*{\fsdurthree} & DICGAN\hspace*{\fsdurthree} & HR\hspace*{\fsdurthree} \\
[4mm]
\multicolumn{8}{c}{201953 from CelebA} \\
\includegraphics[width=0.123\textwidth]{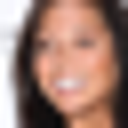}\hspace*{\fsdurthree} &
\includegraphics[width=0.123\textwidth]{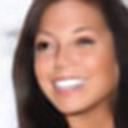}\hspace*{\fsdurthree} &
\includegraphics[width=0.123\textwidth]{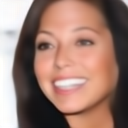}\hspace*{\fsdurthree} &
\includegraphics[width=0.123\textwidth]{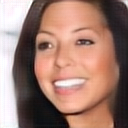}\hspace*{\fsdurthree} &
\includegraphics[width=0.123\textwidth]{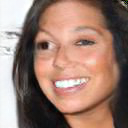}\hspace*{\fsdurthree} &
\includegraphics[width=0.123\textwidth]{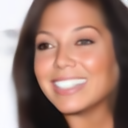}\hspace*{\fsdurthree} &
\includegraphics[width=0.123\textwidth]{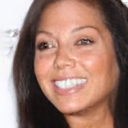}\hspace*{\fsdurthree} &
\includegraphics[width=0.123\textwidth]{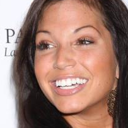}\hspace*{\fsdurthree} \\
Bicubic\hspace*{\fsdurthree} & RDN\hspace*{\fsdurthree} & FSRNet\hspace*{\fsdurthree} & FSRGAN\hspace*{\fsdurthree} & PFSR\hspace*{\fsdurthree} & DIC\hspace*{\fsdurthree} & DICGAN\hspace*{\fsdurthree} & HR\hspace*{\fsdurthree} \\
[4mm]
\multicolumn{8}{c}{3219692565\_1 from Helen} \\
\includegraphics[width=0.123\textwidth]{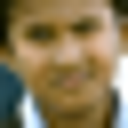}\hspace*{\fsdurthree} &
\includegraphics[width=0.123\textwidth]{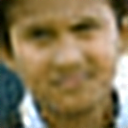}\hspace*{\fsdurthree} &
\includegraphics[width=0.123\textwidth]{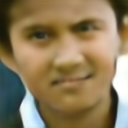}\hspace*{\fsdurthree} &
\includegraphics[width=0.123\textwidth]{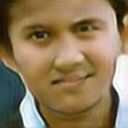}\hspace*{\fsdurthree} &
\includegraphics[width=0.123\textwidth]{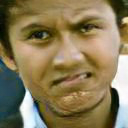}\hspace*{\fsdurthree} &
\includegraphics[width=0.123\textwidth]{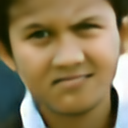}\hspace*{\fsdurthree} &
\includegraphics[width=0.123\textwidth]{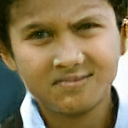}\hspace*{\fsdurthree} &
\includegraphics[width=0.123\textwidth]{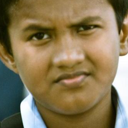}\hspace*{\fsdurthree} \\
Bicubic\hspace*{\fsdurthree} & RDN\hspace*{\fsdurthree} & FSRNet\hspace*{\fsdurthree} & FSRGAN\hspace*{\fsdurthree} & PFSR\hspace*{\fsdurthree} & DIC\hspace*{\fsdurthree} & DICGAN\hspace*{\fsdurthree} & HR\hspace*{\fsdurthree} \\
[4mm]
\multicolumn{8}{c}{3255054809\_1 from Helen} \\
\includegraphics[width=0.123\textwidth]{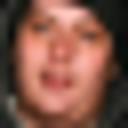}\hspace*{\fsdurthree} &
\includegraphics[width=0.123\textwidth]{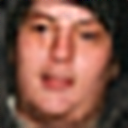}\hspace*{\fsdurthree} &
\includegraphics[width=0.123\textwidth]{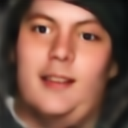}\hspace*{\fsdurthree} &
\includegraphics[width=0.123\textwidth]{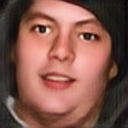}\hspace*{\fsdurthree} &
\includegraphics[width=0.123\textwidth]{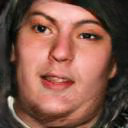}\hspace*{\fsdurthree} &
\includegraphics[width=0.123\textwidth]{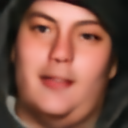}\hspace*{\fsdurthree} &
\includegraphics[width=0.123\textwidth]{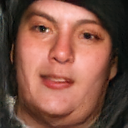}\hspace*{\fsdurthree} &
\includegraphics[width=0.123\textwidth]{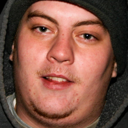}\hspace*{\fsdurthree} \\
Bicubic\hspace*{\fsdurthree} & RDN\hspace*{\fsdurthree} & FSRNet\hspace*{\fsdurthree} & FSRGAN\hspace*{\fsdurthree} & PFSR\hspace*{\fsdurthree} & DIC\hspace*{\fsdurthree} & DICGAN\hspace*{\fsdurthree} & HR\hspace*{\fsdurthree} \\
[4mm]

\end{tabular}
\end{adjustbox}
\vspace{1mm}

\caption{Qualitative comparison with state-of-the-art face super-resolution methods.}
\label{fig:supp_vis2}
\end{figure*}

In Figure~\ref{fig:supp_vis1} and Figure~\ref{fig:supp_vis2} (\textbf{the next pages}), we present more qualitative comparison with state-of-the-art FSR methods including RDN~\cite{zhang2018residual}, FSRNet~\cite{chen2018fsrnet}, FSRGAN~\cite{chen2018fsrnet} and PFSR~\cite{kim2019progressive}. The results demonstrate the effectiveness of our proposed method.

\begin{table}
\centering
\caption{Detailed architecture of the recurrent SR branch. }
\vspace{5px}
\label{tab:SR_branch}
\renewcommand{\arraystretch}{1.2}
  \begin{tabular}{|c|c|}
  \hline
    Layer & Output size \\
    \hline
    \hline
    Input $I^{LR}$ & $16\times 16 \times 3$ \\
    \hline
    Conv ($G_1$) & $16 \times 16 \times 192$ \\
    \hline
    PixelShuffle ($G_1$) & $32 \times 32 \times 48$ \\
    \hline
    Concatenation & $32 \times 32 \times 96$ \\
    \hline
    Conv ($G_R$) & $32 \times 32 \times 48$ \\
    \hline
    Attentive Fusion ($G_R$) & $32 \times 32 \times 48$ \\
    \hline
    Recurrent SR Module ($G_R$) & $32 \times 32 \times 48$ \\
    \hline
    Deconv ($G_2$) & $128 \times 128 \times 48$ \\
    \hline
    Conv ($G_2$) & $128 \times 128 \times 3$ \\
    \hline
    Addition & $128 \times 128 \times 3$ \\
    \hline
    Output $I^{SR}$ & $128 \times 128 \times 3$ \\
    \hline
  \end{tabular}
\vspace{3mm}
\end{table}

\begin{table}
\centering
\caption{Detailed architecture of the recurrent alignment branch. }
\vspace{5px}
\label{tab:Align_branch}
\renewcommand{\arraystretch}{1.2}
  \begin{tabular}{|c|c|}
  \hline
    Layer & Output size \\
    \hline
    \hline
    Input $I^{SR}$ & $128 \times 128 \times 3$ \\
    \hline
    $A_1$ & $32 \times 32 \times 256$ \\
    \hline
    Concatenation & $32 \times 32 \times 512$ \\
    \hline
    Conv ($A_R$) & $32 \times 32 \times 512$ \\
    \hline
    Recurrent HourGlass ($A_R$) & $32 \times 32 \times 512$ \\
    \hline
    \multirow{2}{*}{Split} & $32 \times 32 \times 256$ \\
    \cline{2-2}
    & $32 \times 32 \times 256$ \\
    \hline
    $A_2$ & $32 \times 32 \times 68$ \\
    \hline
    Output $L$ & $32 \times 32 \times 68$ \\
    \hline
  \end{tabular}
\end{table}

\end{document}